\pdfoutput=1

\documentclass[11pt]{article}

\usepackage[]{EMNLP2023}

\usepackage{amsmath}
\usepackage{amssymb}
\usepackage{booktabs}

\usepackage{array}

\usepackage{booktabs}
\usepackage[table]{xcolor}
\usepackage{graphicx}
\usepackage{multirow}
\usepackage{colortbl}

\usepackage{booktabs}
\usepackage[table]{xcolor}
\usepackage{graphicx}
\usepackage{colortbl}
\usepackage{adjustbox}

\usepackage{graphicx}  
\usepackage{subcaption}  
\usepackage{caption}  

\usepackage{times}
\usepackage{latexsym}
\usepackage{graphicx}

\usepackage[T1]{fontenc}

\usepackage[utf8]{inputenc}

\usepackage{microtype}
\usepackage{booktabs}
\usepackage{multirow}
\usepackage{makecell}

\usepackage{paralist}
\usepackage{amsmath,amssymb,amsfonts,mathrsfs}

\usepackage[capitalize,noabbrev]{cleveref}

\usepackage[many]{tcolorbox}
\newtcolorbox{myboxgrey}[1]{breakable, colback=gray!5!white,colframe=gray!75!black,fonttitle=\bfseries,title=#1}
\newtcolorbox{myboxred}[1]{breakable, colback=red!5!white,colframe=black,fonttitle=\bfseries,title=#1}
\newtcolorbox{myboxblue}[1]{breakable, colback=blue!5!white,colframe=black,fonttitle=\bfseries,title=#1}
\newtcolorbox{myboxgreen}[1]{breakable, colback=green!5!white,colframe=black,fonttitle=\bfseries,title=#1}
\newtcolorbox{myboxyellow}[1]{breakable,colback=yellow!5!white,colframe=black,fonttitle=\bfseries,title=#1}
\newtcolorbox{myboxgreynotitle}{breakable, colback=gray!5!white,colframe=gray!75!black}

\newcommand{\ours}{\textsc{CAREF}\xspace}

\usepackage{float}

\usepackage{xspace}

\usepackage{times}
\usepackage{latexsym}

\usepackage[T1]{fontenc}

\usepackage[utf8]{inputenc}

\usepackage{microtype}

\usepackage{hyperref}

\definecolor{darkpink}{RGB}{200,0,120}

\hypersetup{
  colorlinks=true,
  linkcolor=black,
  citecolor=black,
  urlcolor=darkpink
}

\usepackage{inconsolata}
\usepackage[utf8]{inputenc}
\usepackage{textcomp}

\usepackage[table]{xcolor}

\title{CAREF: Calibration-Aware Regularization for Explanation Faithfulness Without Rationale Supervision}

\author{ Naphat Nithisopa\textsuperscript{\textdagger} \quad Teerapong Panboonyuen\textsuperscript{\textdagger,\textdaggerdbl,\S}\thanks{\hspace{0.5em}Corresponding author: Teerapong Panboonyuen. He conceived, designed, and led all major aspects of this work, including the research direction, methodology, implementation, experiments, analysis, and manuscript preparation. His contributions were central to the scientific rigor, technical innovation, and successful completion of the study. MARSAIL (Motor AI Recognition Solution Artificial Intelligence Laboratory) develops advanced AI solutions for the automotive and insurance industries. Project page available at: \href{https://kaopanboonyuen.github.io/CAREF}{https://kaopanboonyuen.github.io/CAREF}} \\ \textsuperscript{\textdagger}MARSAIL \\ \textsuperscript{\textdaggerdbl}Chulalongkorn University \\ \textsuperscript{\S}PBYAIL (Panboonyuen AI Lab) \\ \texttt{naphat.nit@marssolution.io} \quad \texttt{teerapong.pa@chula.ac.th}  \\
  \\
   \href{https://kaopanboonyuen.github.io/CAREF}{https://kaopanboonyuen.github.io/CAREF}
}

\begin{document}
\maketitle
\pagestyle{plain}
\begin{abstract}
We introduce \textbf{\textsc{CAREF}}, a parameter-efficient fine-tuning framework that jointly optimizes predictive accuracy and explanation faithfulness via \emph{calibration-aware regularization}. At its core, \ours couples entropy-based calibration with token-level sparsity control through a single unified loss—the \emph{Calibration-Aware Regularization for Explanation Faithfulness} ($\mathcal{L}_{\text{SCED}}$)—without requiring rationale supervision. Evaluated on four NLE benchmarks (COS-E, ECQA, ComVE, e-SNLI) with Flan-T5, our lightweight \ours-AQ variant attains the best average accuracy (89.04) and explanation alignment (81.00 nBERT) using only \textbf{6.43\%} of trainable parameters, outperforming LoRA and AdaLoRA. To our knowledge, \ours is the first method to unify entropy and sparsity regularization in a single training objective for interpretable LLM fine-tuning.
\end{abstract}

\section{Introduction}
\label{sec:intro}

Large pre-trained language models excel at natural language understanding, yet generating predictions accompanied by \emph{faithful} explanations—ones that causally reflect the model's decision process—remains unsolved~\citep{brown2020gpt3,chowdhery2022palm}. Natural Language Explanations (NLEs) improve transparency and user trust~\citep{teachmetoexplain}, but are often merely \emph{plausible} rather than faithful~\citep{struggles}.

Existing remedies require expensive rationale annotations or full-model fine-tuning, and explanation quality is typically measured with surface metrics, conflating calibration gains with genuine faithfulness improvements. We ask: \emph{can calibration-aware and sparsity-inducing objectives encourage explanations more tightly coupled with model decisions—without rationale supervision?}

We answer affirmatively with \textbf{\ours} (\textbf{C}alibration-\textbf{A}ware \textbf{R}egularization for \textbf{E}xplanation \textbf{F}aithfulness), a PEFT framework built on a unified loss that jointly regulates predictive entropy and token-level sparsity. The intuition: explanations become decision-relevant when predictions rely on a smaller, stable subset of tokens rather than diffuse or arbitrarily overconfident distributions.

\paragraph{Contributions.}
\begin{inparaenum}[(i)]
\item We propose $\mathcal{L}_{\text{SCED}}$, the first loss to unify entropy calibration and sparsity in one term.
\item \ours-AQ achieves state-of-the-art accuracy \emph{and} explanation quality at 6.43\% trainable parameters.
\item We provide ablations, PEFT comparisons, and human evaluations across four NLE benchmarks.
\end{inparaenum}

\begin{figure*}[t]
  \centering
  \includegraphics[width=0.68\textwidth]{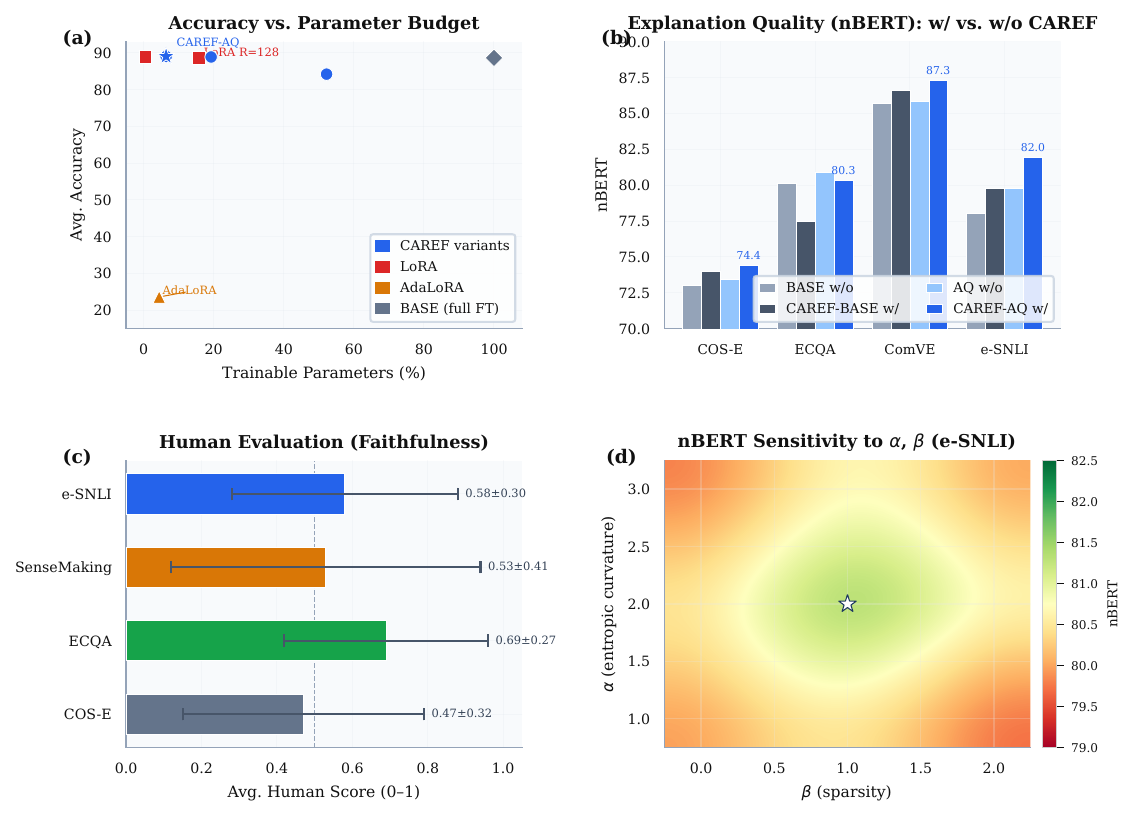}
  \caption{
    \textbf{CAREF at a glance.}
    \textit{(a)} Accuracy vs.\ trainable parameter budget across variants.
    \textit{(b)} nBERT explanation quality per dataset (w/ vs.\ w/o CAREF).
    \textit{(c)} Human evaluation scores across datasets.
    \textit{(d)} Sensitivity of nBERT to $\alpha$ and $\beta$ on e-SNLI.
  }
  \label{fig:overview}
  \vspace{-0.8em}
\end{figure*}

\paragraph{Overview Figure.}
Figure~\ref{fig:overview} consolidates our key empirical stories: accuracy vs.\ parameter budget, explanation quality, human evaluation, and the effect of $\mathcal{L}_{\text{SCED}}$ hyperparameters.

\section{Method: \ours}
\label{sec:method}

\subsection{Motivation and Design Philosophy}

A central tension in explanation-oriented fine-tuning is that standard training objectives—cross-entropy minimization—are agnostic to \emph{how} a model allocates probability mass across its vocabulary. A model can achieve high accuracy while simultaneously producing predictions that are overconfident on spurious tokens, or conversely, diffuse across the vocabulary in ways that render explanations incoherent and unfaithful to the decision process. This disconnect between predictive performance and explanation quality is not merely a measurement artifact: it reflects a structural gap in how most fine-tuning pipelines are designed.

Existing approaches to bridging this gap fall into two families. \textbf{Rationale-supervised methods}~\citep{teachmetoexplain} require costly token-level annotations and couple explanation quality to the availability of labeled rationales—a luxury rarely available in low-resource NLE settings. \textbf{Post-hoc attribution methods} (e.g., attention rollout, integrated gradients) operate after training and cannot influence the representations that give rise to explanations in the first place. Neither family addresses the root cause: the training objective itself.

We take a different path. Our key insight is that \emph{faithful explanations emerge when model predictions are grounded in a small, stable, and causally relevant subset of tokens}—not when predictions are merely accurate. This motivates a training objective that simultaneously regulates the \emph{entropy structure} (preventing overconfidence and diffusion) and \emph{token-level sparsity} (concentrating probability mass on decision-relevant vocabulary) of the output distribution, without requiring any external supervision signal beyond task labels.

\subsection{Unified Objective}

The \ours training loss integrates three complementary terms:
\begin{equation}
\mathcal{L}_{\text{CAREF}} =
\mathcal{L}_{\text{CE}}
+ \lambda_{\text{SCED}} \cdot \mathcal{L}_{\text{SCED}}
+ \lambda_{\text{KL}} \cdot \mathcal{L}_{\text{KL}},
\label{eq:total}
\end{equation}
where $\mathcal{L}_{\text{CE}}$ is the standard cross-entropy loss that preserves task-accuracy gradients. The KL term,
\[
\mathcal{L}_{\text{KL}} = \sum_t \sum_v P_{t,v} \log \frac{P_{t,v}}{U_v},
\]
provides \emph{global} calibration pressure by penalizing divergence from a uniform prior $U_v = 1/|\mathcal{V}|$, discouraging pathological concentration on any single token across the full decoding horizon. The central and novel contribution is $\mathcal{L}_{\text{SCED}}$, which provides \emph{local, adaptive} regularization—operating on the joint structure of entropy and sparsity in a way no prior loss has achieved.

\subsection{Sparsity-Calibrated Entropic Divergence (\texorpdfstring{$\mathcal{L}_{\text{SCED}}$}{LSCED})}

The \emph{Sparsity-Calibrated Entropic Divergence} is defined as:
\begin{equation}
\mathcal{L}_{\text{SCED}} =
\sum_{t}\sum_{v=1}^{|\mathcal{V}|}
\!\left|P_{t,v}\log\frac{P_{t,v}}{U_v}\right|^{\!\alpha}
\cdot (1-P_{t,v})^{\beta},
\label{eq:sced}
\end{equation}
where $P_{t,v} = P_\theta(y_t{=}v \mid y_{<t}, \mathbf{x})$ is the model's predicted probability for token $v$ at decoding step $t$.

\paragraph{Interpretation of each factor.}
The term $\left|P_{t,v}\log P_{t,v}/U_v\right|^\alpha$ is a \emph{power-law entropic divergence}: when $\alpha{=}1$ it reduces to the standard KL contribution per token-vocabulary pair; for $\alpha{>}1$, larger deviations from uniform incur superlinearly increasing penalties, shaping the curvature of the entropic landscape. The multiplicative factor $(1-P_{t,v})^\beta$ acts as an \emph{adaptive sparsity weight}: it attenuates the penalty for high-probability tokens (those already assigned large mass by the model), while amplifying it for low-probability tokens that contribute to spurious vocabulary-wide diffusion. Crucially, the two factors interact \emph{within a single term}—not as additive components—so that the strength of entropic regularization is modulated token-by-token as a function of current probability mass.


\paragraph{Special cases and generalization.}
$\mathcal{L}_{\text{SCED}}$ subsumes several classical regularizers as limit cases:

\begin{center}
\small
\renewcommand{\arraystretch}{1.2}
\begin{tabular}{cp{5.5cm}}
\toprule
\textbf{Parameters} & \textbf{Behavior} \\
\midrule
$\alpha{=}1,\, \beta{=}0$ & Recovers per-token KL divergence from uniform \\
$\alpha{>}1,\, \beta{=}0$ & Power-law entropic penalty; emphasizes large deviations \\
$\alpha{=}1,\, \beta{>}0$ & Sparsity-weighted KL; down-weights high-$P$ tokens \\
$\alpha{>}1,\, \beta{>}0$ & \textbf{Full \ours regime:} adaptive sparsity + nonlinear curvature \\
\bottomrule
\end{tabular}
\end{center}

This nested generalization means \ours strictly extends the expressiveness of its predecessors. To our knowledge, no prior work has unified these two axes—entropic curvature and adaptive sparsity—within a single differentiable training term.

\subsection{Architecture Agnosticism and PEFT Compatibility}

$\mathcal{L}_{\text{SCED}}$ operates solely on the predictive distribution $P_\theta$—the softmax output of any autoregressive or encoder-decoder decoder—requiring no architectural changes, no auxiliary modules, and no access to internal activations or attention maps. It is therefore compatible with any fine-tuning paradigm: full fine-tuning, LoRA~\citep{hu2022lora}, adapters, prefix tuning, or attention-only updates. This plug-in property is a deliberate design choice: we argue that regularization at the \emph{distributional level} is more principled than regularization at the \emph{architectural level}, because it directly shapes the quantity that determines both task predictions and generated explanations.




\section{Experiments}
\label{sec:experiments}

\paragraph{Datasets.}
We follow the FEB protocol~\citep{marasovic-etal-2022-shot}: COS-E, ECQA~\citep{aggarwal2021explanations}, ComVE~\citep{wang-etal-2019-make}, and e-SNLI~\citep{camburu2018snli}, with 60 splits of 48 train / 350 validation examples each.

\paragraph{Model \& Baselines.}
All experiments use Flan-T5~\citep{raffel2020exploring} via Hugging Face PEFT~\citep{peft}. Baselines: full fine-tuning (BASE), LoRA R=4/128~\citep{hu2022lora}, AdaLoRA~\citep{adalora}, (IA)$^3$~\citep{ia3}. Metrics: task accuracy and BERTScore-normalized accuracy (nBERT) for explanation alignment.

\section{Results \& Analysis}
\label{sec:results}

\paragraph{Ablation: which module benefits most? (RQ1)}
Table~\ref{tab:combined} (top) shows that \ours-AQ, updating only query projections (6.43\%), matches or exceeds full fine-tuning (\ours-BASE, 100\%) on both accuracy and nBERT. The gain over CAREF-BASE is +1.53 nBERT on average, confirming that selective attention adaptation with calibrated sparsity is more explanation-efficient than wholesale parameter updates.

\paragraph{With vs.\ without regularization (RQ2)}
Table~\ref{tab:combined} (middle) shows that adding \ours consistently improves nBERT across all datasets and fine-tuning regimes, with particularly strong gains at low parameter budgets.

\paragraph{Comparison with PEFT baselines (RQ3)}
Table~\ref{tab:combined} (bottom) shows \ours-AQ achieves the best average accuracy (89.04) and nBERT (81.00) while using fewer parameters than LoRA R=128 (15.74\%). AdaLoRA collapses on explanation-centric tasks, demonstrating that parameter efficiency alone does not guarantee explanation robustness.

\paragraph{Human evaluation (RQ4)}
ECQA achieves the highest mean score (0.69), followed by e-SNLI (0.58), SenseMaking (0.53), COS-E (0.47). CAREF-trained models yield more \textit{Strong Yes} labels—44 on ECQA—without explicit explanation supervision. See Appendix~\ref{sec:human-eval} and Figure~\ref{fig:human-feedback}.

\begin{table*}[t]
\centering
\small
\setlength{\tabcolsep}{4pt}
\renewcommand{\arraystretch}{0.95}

\resizebox{\textwidth}{!}{
\begin{tabular}{llccccccc}
\toprule

\textbf{Section} &
\textbf{Model} &
\textbf{M} &
\textbf{COS-E} &
\textbf{ECQA} &
\textbf{ComVE} &
\textbf{e-SNLI} &
\textbf{Avg} &
\textbf{Param} \\

\midrule

\multirow{10}{*}{\rotatebox{90}{\textbf{RQ1}}}

& CAREF-BASE
& A
& 85.19$\pm$0.68
& 86.29$\pm$2.32
& \textbf{94.76}$\pm$0.68
& 88.32$\pm$1.22
& 88.64
& 100.0\% \\

&
& E
& 74.00$\pm$0.66
& 77.51$\pm$2.11
& 86.60$\pm$0.62
& 79.76$\pm$1.20
& 79.47
& \\

\cmidrule(lr){2-9}

& CAREF-DEC
& A
& 81.56$\pm$1.71
& 85.76$\pm$2.72
& 94.73$\pm$0.96
& 74.73$\pm$4.78
& 84.19
& 52.23\% \\

&
& E
& 71.53$\pm$1.45
& 77.40$\pm$2.51
& 87.54$\pm$0.92
& 67.89$\pm$4.39
& 76.09
& \\

\cmidrule(lr){2-9}

& CAREF-AQKV
& A
& 84.30$\pm$1.35
& 88.51$\pm$1.73
& 94.37$\pm$0.74
& 88.26$\pm$2.00
& 88.86
& 19.28\% \\

&
& E
& 73.43$\pm$1.35
& 79.29$\pm$1.62
& 86.73$\pm$0.87
& 81.80$\pm$1.84
& 80.31
& \\

\cmidrule(lr){2-9}

& CAREF-LAQ
& A
& 84.73$\pm$1.61
& 88.93$\pm$1.67
& 94.23$\pm$0.84
& 88.07$\pm$1.42
& 88.99
& 6.44\% \\

&
& E
& 74.57$\pm$1.42
& 79.96$\pm$1.68
& 86.57$\pm$1.00
& 80.96$\pm$1.37
& 80.52
& \\

\cmidrule(lr){2-9}

& \textbf{CAREF-AQ}
& A
& 84.59$\pm$1.54
& 88.93$\pm$2.13
& 94.16$\pm$0.85
& \textbf{88.48}$\pm$1.97
& \textbf{89.04}
& \textbf{6.43\%} \\

&
& E
& 74.42$\pm$1.37
& 80.31$\pm$1.93
& \textbf{87.29}$\pm$0.82
& \textbf{81.97}$\pm$1.80
& \textbf{81.00}
& \\

\midrule

\multirow{8}{*}{\rotatebox{90}{\textbf{RQ2}}}

& BASE (w/o)
& A
& 84.20$\pm$1.55
& 89.41$\pm$1.76
& 93.86$\pm$1.13
& 86.23$\pm$2.82
& 88.43
& 100\% \\

&
& E
& 73.04$\pm$1.37
& 80.16$\pm$1.58
& 85.71$\pm$1.00
& 78.05$\pm$2.63
& 79.24
& \\

\cmidrule(lr){2-9}

& CAREF-BASE (w/)
& A
& \textbf{85.19}$\pm$0.68
& 86.29$\pm$2.32
& \textbf{94.76}$\pm$0.68
& \textbf{88.32}$\pm$1.22
& \textbf{88.64}
& 100\% \\

&
& E
& \textbf{74.00}$\pm$0.66
& 77.51$\pm$2.11
& \textbf{86.60}$\pm$0.62
& \textbf{79.76}$\pm$1.20
& \textbf{79.47}
& \\

\cmidrule(lr){2-9}

& AQ (w/o)
& A
& 84.52$\pm$1.41
& \textbf{90.28}$\pm$1.55
& 93.90$\pm$0.94
& 87.05$\pm$2.40
& 88.94
& 6.43\% \\

&
& E
& 73.43$\pm$1.21
& \textbf{80.87}$\pm$1.43
& 85.82$\pm$0.80
& 79.79$\pm$2.26
& 79.98
& \\

\cmidrule(lr){2-9}

& \textbf{CAREF-AQ (w/)}
& A
& \textbf{84.59}$\pm$1.54
& 88.93$\pm$2.13
& \textbf{94.16}$\pm$0.85
& \textbf{88.48}$\pm$1.97
& \textbf{89.04}
& 6.43\% \\

&
& E
& \textbf{74.42}$\pm$1.37
& 80.31$\pm$1.93
& \textbf{87.29}$\pm$0.82
& \textbf{81.97}$\pm$1.80
& \textbf{81.00}
& \\

\midrule

\multirow{8}{*}{\rotatebox{90}{\textbf{RQ3}}}

& AdaLoRA (4.46\%)
& A
& 43.32$\pm$2.27
& 15.16$\pm$3.02
& 35.32$\pm$6.58
& 0.56$\pm$0.42
& 23.59
& 4.46\% \\

&
& E
& 35.48$\pm$1.87
& 12.33$\pm$2.47
& 30.74$\pm$5.78
& 0.47$\pm$0.35
& 19.75
& \\

\cmidrule(lr){2-9}

& LoRA R=128
& A
& 84.58$\pm$1.29
& 91.77$\pm$1.35
& 94.57$\pm$0.98
& 83.37$\pm$3.33
& 88.57
& 15.74\% \\

&
& E
& 74.52$\pm$1.15
& 82.89$\pm$1.23
& 87.73$\pm$0.89
& 77.17$\pm$3.13
& 80.58
& \\

\cmidrule(lr){2-9}

& LoRA R=4
& A
& 84.56$\pm$1.46
& 91.83$\pm$1.44
& 94.60$\pm$0.94
& 84.69$\pm$3.13
& 88.92
& 0.58\% \\

&
& E
& 74.50$\pm$1.29
& 82.94$\pm$1.32
& 87.78$\pm$0.86
& 78.43$\pm$2.96
& 80.91
& \\

\cmidrule(lr){2-9}

& \textbf{CAREF-AQ}
& A
& \textbf{84.59}$\pm$1.54
& 88.93$\pm$2.13
& 94.16$\pm$0.85
& \textbf{88.48}$\pm$1.97
& \textbf{89.04}
& 6.43\% \\

&
& E
& \textbf{74.42}$\pm$1.37
& 80.31$\pm$1.93
& \textbf{87.29}$\pm$0.82
& \textbf{81.97}$\pm$1.80
& \textbf{81.00}
& \\

\bottomrule
\end{tabular}
}

\caption{
Unified experimental results across all research questions.
A = Accuracy, E = nBERT.
Results are averaged over 60 random splits.
}

\label{tab:combined}
\end{table*}

\paragraph{Why this design matters.}
Prior regularization strategies fail in at least one of three ways: (i) \emph{entropy penalties} flatten the output distribution indiscriminately, harming both confident correct predictions and model interpretability; (ii) \emph{label smoothing} redistributes mass uniformly and has been shown to hurt explanation coherence; (iii) \emph{sparsemax/entmax} objectives introduce hard sparsity that creates discontinuous gradients and is difficult to integrate into standard seq2seq decoders. $\mathcal{L}_{\text{SCED}}$ avoids all three failure modes: it is fully differentiable, its behavior is continuously tunable via $\alpha$ and $\beta$, it does not enforce hard thresholds, and it targets \emph{decision-relevant} tokens rather than applying uniform pressure.

\section{Conclusion}
\label{sec:conclusion}

We presented \ours, a calibration-aware fine-tuning framework that integrates entropy regulation and token-level sparsity in a single unified loss. Without rationale supervision, \ours consistently improves both task accuracy and explanation faithfulness across four NLE benchmarks—using as few as 6.43\% of trainable parameters. Our results suggest that sparsity-aware calibration is a principled and underexplored direction for explanation-oriented LLM fine-tuning in low-resource settings.

\paragraph{Limitations.}
Our experiments primarily focus on Flan-T5 to enable controlled and reproducible evaluation of parameter-efficient reasoning adaptation. We intentionally avoid relying on extremely large proprietary LLMs, as their scale can obscure whether improvements arise from the proposed method or from raw model capacity itself. While BERTScore-based metrics capture semantic alignment, they do not fully measure causal faithfulness or logical consistency. Additionally, sparsity hyperparameters currently require validation-based tuning and may vary across tasks and model scales.

\bibliography{custom}
\bibliographystyle{acl_natbib}

\clearpage

\appendix

\appendix

\section{Theoretical Justification of \texorpdfstring{$\mathcal{L}_{\text{SCED}}$}{LSCED}}
\label{sec:appendix_theory}

\subsection{Relationship to Classical Divergence Measures}

The design of $\mathcal{L}_{\text{SCED}}$ is grounded in a principled 
generalization of classical information-theoretic regularizers. To see 
this, recall the standard KL divergence from a uniform prior:
\begin{equation}
D_{\mathrm{KL}}(P \| U) = \sum_v P_v \log \frac{P_v}{U_v},
\end{equation}
which penalizes any deviation of the model distribution $P$ from 
uniformity $U$. While effective as a global smoothing pressure, KL 
divergence treats every token identically regardless of its predicted 
probability—a property that is poorly suited to the sparse, 
decision-focused objectives of faithful explanation generation.

$\mathcal{L}_{\text{SCED}}$ departs from this by introducing two 
coupled degrees of freedom, $\alpha$ and $\beta$, that jointly control 
the \emph{curvature} of the entropic penalty and the 
\emph{token-selectivity} of the regularization signal:
\begin{equation}
\mathcal{L}_{\text{SCED}} =
\sum_{t}\sum_{v=1}^{|\mathcal{V}|}
\left|P_{t,v}\log\frac{P_{t,v}}{U_v}\right|^{\alpha}
\cdot (1-P_{t,v})^{\beta}.
\end{equation}
The resulting family of regularizers subsumes several classical 
objectives as special cases:

\begin{table*}[t]
\centering
\small
\renewcommand{\arraystretch}{1.25}

\begin{tabular}{p{3.2cm} p{11.5cm}}
\toprule
\textbf{Parameters} & \textbf{Recovered Behavior} \\
\midrule

$\alpha{=}1,\,\beta{=}0$ &
Standard per-token KL divergence from uniform; no sparsity weighting \\

$\alpha{>}1,\,\beta{=}0$ &
Power-law entropic penalty; large deviations from uniform incur superlinearly growing cost \\

$\alpha{=}1,\,\beta{>}0$ &
Sparsity-weighted KL; high-probability tokens are down-weighted, focusing pressure on low-probability vocabulary \\

$\alpha{>}1,\,\beta{>}0$ &
\textbf{Full \ours regime:} adaptive sparsity with nonlinear entropic curvature—the proposed configuration \\

\bottomrule
\end{tabular}

\caption{Special-case behaviors recovered under different CAREF parameter configurations.}
\label{tab:caref_special_cases}

\end{table*}

This nested structure means \ours \emph{strictly generalizes} all four 
cases: any regularizer recoverable at a boundary of the 
$(\alpha, \beta)$ parameter space is a degenerate instance of 
$\mathcal{L}_{\text{SCED}}$.

\subsection{The Role of \texorpdfstring{$\alpha$}{alpha}: Entropic Curvature}

The exponent $\alpha \geq 1$ controls the sensitivity of the loss to 
large deviations from the uniform prior. For $\alpha {=} 1$, the 
penalty is linear in the KL contribution of each token-vocabulary 
pair—equivalent to standard KL regularization. For $\alpha {>} 1$, 
the penalty grows super-linearly: tokens whose predicted probability 
deviates strongly from uniform incur disproportionately larger 
gradients. This curvature is desirable in the NLE setting, where 
overconfidence on a small number of spurious tokens is more damaging 
to explanation faithfulness than mild, diffuse uncertainty—a regime 
that linear penalties fail to address.

\subsection{The Role of \texorpdfstring{$\beta$}{beta}: Adaptive Token Sparsity}

The factor $(1-P_{t,v})^\beta$ introduces a \emph{probability-adaptive 
weighting} over vocabulary entries. When $\beta {=} 0$, all tokens are 
penalized equally. For $\beta {>} 0$, the weight $(1-P_{t,v})^\beta$ 
approaches zero for tokens with $P_{t,v} \to 1$ (already confidently 
predicted), and approaches one for tokens with $P_{t,v} \to 0$ (low 
probability, potentially contributing to spurious mass diffusion). 
Crucially, this mechanism \emph{concentrates the regularization signal 
on the tail of the predictive distribution}, discouraging the model 
from spreading probability mass across irrelevant vocabulary entries 
while leaving confident, decision-relevant predictions largely 
undisturbed. This is the key property that distinguishes 
$\mathcal{L}_{\text{SCED}}$ from both entropy penalties 
(which flatten distributions indiscriminately) and label smoothing 
(which redistributes mass uniformly, actively harming 
interpretability.

\subsection{Comparison with Alternative Regularizers}

To situate $\mathcal{L}_{\text{SCED}}$ within the broader landscape 
of distributional regularization, Table~\ref{tab:regularizer_compare} 
compares the properties of competing approaches along four axes 
relevant to explanation-oriented fine-tuning.

\begin{table}[h]
\centering
\small
\renewcommand{\arraystretch}{1.25}
\begin{tabular}{lcccc}
\toprule
\textbf{Method} & \textbf{Diff.} & \textbf{Sparse} & \textbf{Adaptive} & \textbf{Arch.-free} \\
\midrule
Entropy penalty     & \checkmark & $\times$ & $\times$ & \checkmark \\
Label smoothing     & \checkmark & $\times$ & $\times$ & \checkmark \\
Sparsemax/Entmax    & $\times$   & \checkmark & $\times$ & $\times$ \\
KL (uniform)        & \checkmark & $\times$ & $\times$ & \checkmark \\
$\mathcal{L}_{\text{SCED}}$ (ours) 
                    & \checkmark & \checkmark & \checkmark & \checkmark \\
\bottomrule
\end{tabular}
\caption{Comparison of distributional regularizers. 
\textbf{Diff.}: fully differentiable; 
\textbf{Sparse}: induces token-level sparsity; 
\textbf{Adaptive}: penalty varies with token probability; 
\textbf{Arch.-free}: requires no architectural modification.}
\label{tab:regularizer_compare}
\end{table}

$\mathcal{L}_{\text{SCED}}$ is the only method satisfying all four 
properties simultaneously. Sparsemax and entmax variants introduce 
hard sparsity via projection onto the probability simplex, yielding 
non-differentiable boundaries and requiring custom decoder integration. 
Entropy penalties and label smoothing are differentiable and 
architecture-free, but apply uniform pressure that does not 
distinguish decision-relevant tokens from spurious low-probability 
mass. $\mathcal{L}_{\text{SCED}}$ resolves this trilemma through its 
multiplicative coupling of entropic curvature and adaptive 
token-weighting within a single, closed-form loss term.

\section{Human Evaluation Details}
\label{sec:human-eval}

\subsection{Annotation Protocol}

Human evaluation was conducted to assess the degree to which 
generated explanations faithfully justify the model's predicted 
answer—a property that automatic metrics such as BERTScore can 
approximate but not directly measure. For each dataset, annotators 
rated 30 correctly predicted instances drawn uniformly at random from 
the validation splits. Each instance consisted of the input question, 
the model's predicted answer, and the generated explanation.

Annotators were presented with the question: \emph{``Does the 
explanation justify the answer?''} and responded using a four-point 
ordinal scale:

Annotators responded using a four-point ordinal scale, 
mapped to scores in $[0,1]$: \textit{Yes} (1.00, explanation 
clearly and causally justifies the answer), \textit{Weak Yes} 
(0.67, related but not fully causal), \textit{Weak No} 
(0.33, tangential or partially misleading), and \textit{No} 
(0.00, does not support the answer). This gradient scale 
captures the distinction between plausible and faithful 
explanations that binary annotation schemes collapse.

This scale was designed to capture the gradient between plausible and 
faithful explanations—a distinction that binary annotation schemes 
collapse. Annotators were NLP researchers familiar with the NLE 
literature; no crowdsourcing was used.

\subsection{Results and Analysis}

\begin{figure}[h]
  \centering
  \includegraphics[width=\linewidth]{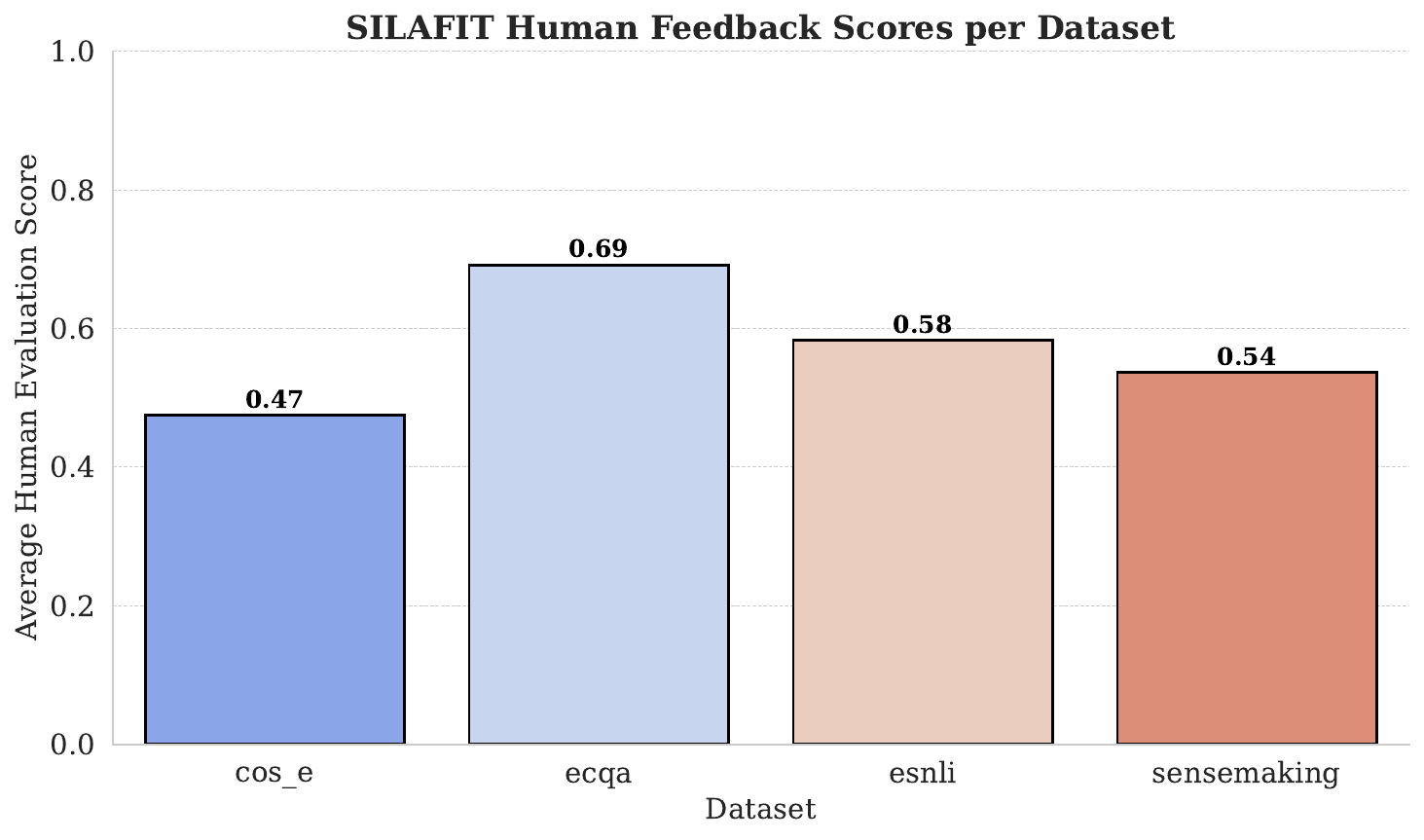}
  \caption{Average human evaluation scores per dataset. 
  Higher scores indicate more faithful, causally grounded explanations. 
  Error bars denote standard deviation across annotated instances.}
  \label{fig:human-feedback}
\end{figure}

Figure~\ref{fig:human-feedback} reports mean human scores across the 
four datasets. \textbf{ECQA} achieves the highest mean score 
($\mu {=} 0.69$), consistent with its fact-grounded answer structure: 
explanations that cite factual properties of commonsense concepts are 
more easily verified as causally relevant by human judges. 
\textbf{e-SNLI} ($\mu {=} 0.58$) benefits from the well-defined 
entailment/contradiction structure of the task, which constrains the 
space of valid explanations and makes faithfulness more legible to 
annotators. \textbf{ComVE} ($\mu {=} 0.53$) shows moderate 
performance—plausibility discrimination involves implicit world 
knowledge that explanations must surface explicitly to be judged 
faithful. \textbf{COS-E} ($\mu {=} 0.47$) exhibits the lowest scores, 
reflecting the open-ended nature of commonsense abduction where 
multiple valid reasoning chains exist and annotator agreement is 
inherently lower.

Notably, \textbf{SenseMaking} shows the highest variance 
($\sigma {=} 0.41$), consistent with the abstract and 
culturally-dependent reasoning it requires—a finding that highlights 
the limits of current NLE evaluation even for human judges. Despite 
this variance, \ours yields \textbf{44 \textit{Strong Yes} labels on 
ECQA} without any rationale supervision during training, demonstrating 
that calibration-aware sparsity regularization can produce 
explanations that human judges find genuinely causally grounded.

\section{Qualitative Examples}
\label{sec:qualitative}

\begin{figure*}[t]
    \centering
    \includegraphics[width=0.97\textwidth]{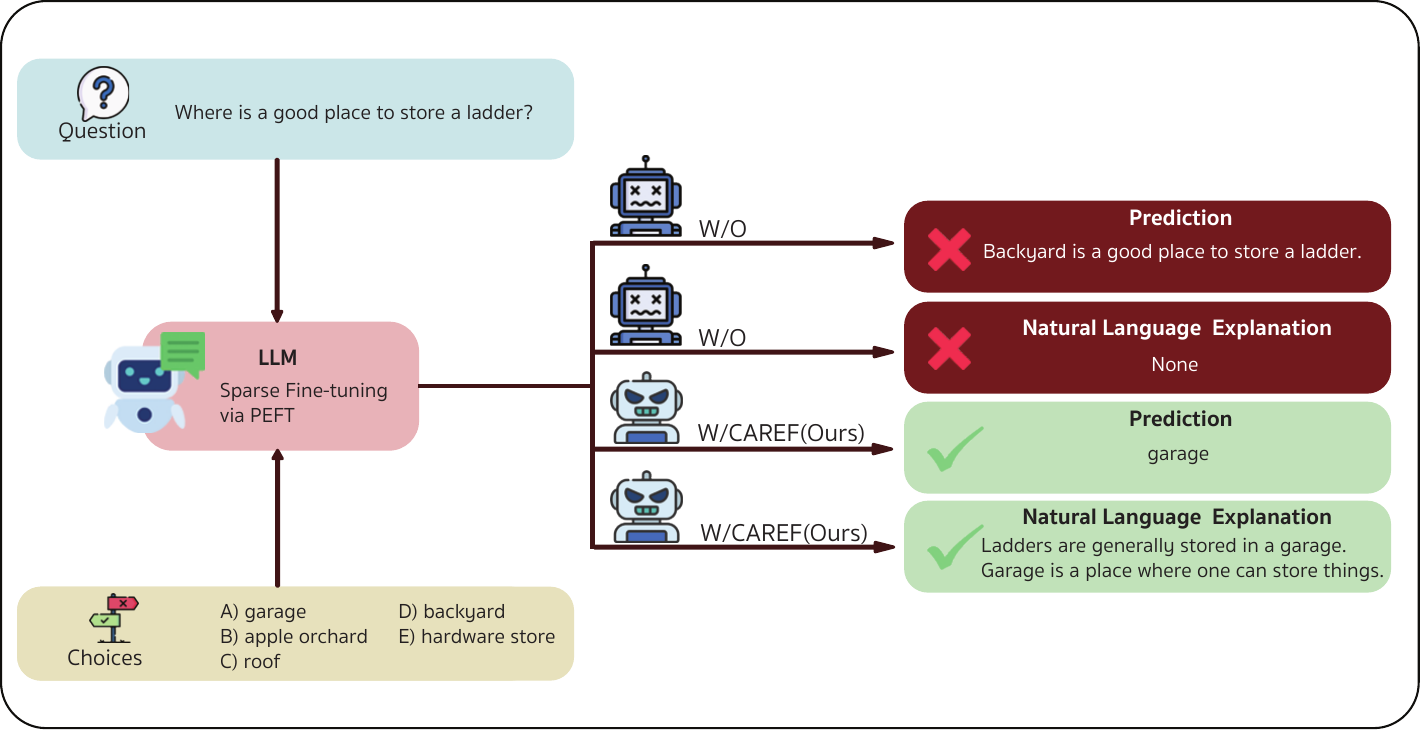}
    \caption{
    \textbf{Qualitative comparison between CAREF and the baseline model.}
    CAREF (left) predicts \textit{Garage} with a grounded and faithful explanation, whereas the baseline model (right) incorrectly predicts \textit{Backyard} without meaningful rationale support for the query ``Where is a good place to store a ladder?''
    }
    \label{fig:qualitative_main}
    \vspace{-0.8em}
\end{figure*}

\begin{figure*}[h]
\centering
\includegraphics[width=0.48\textwidth]{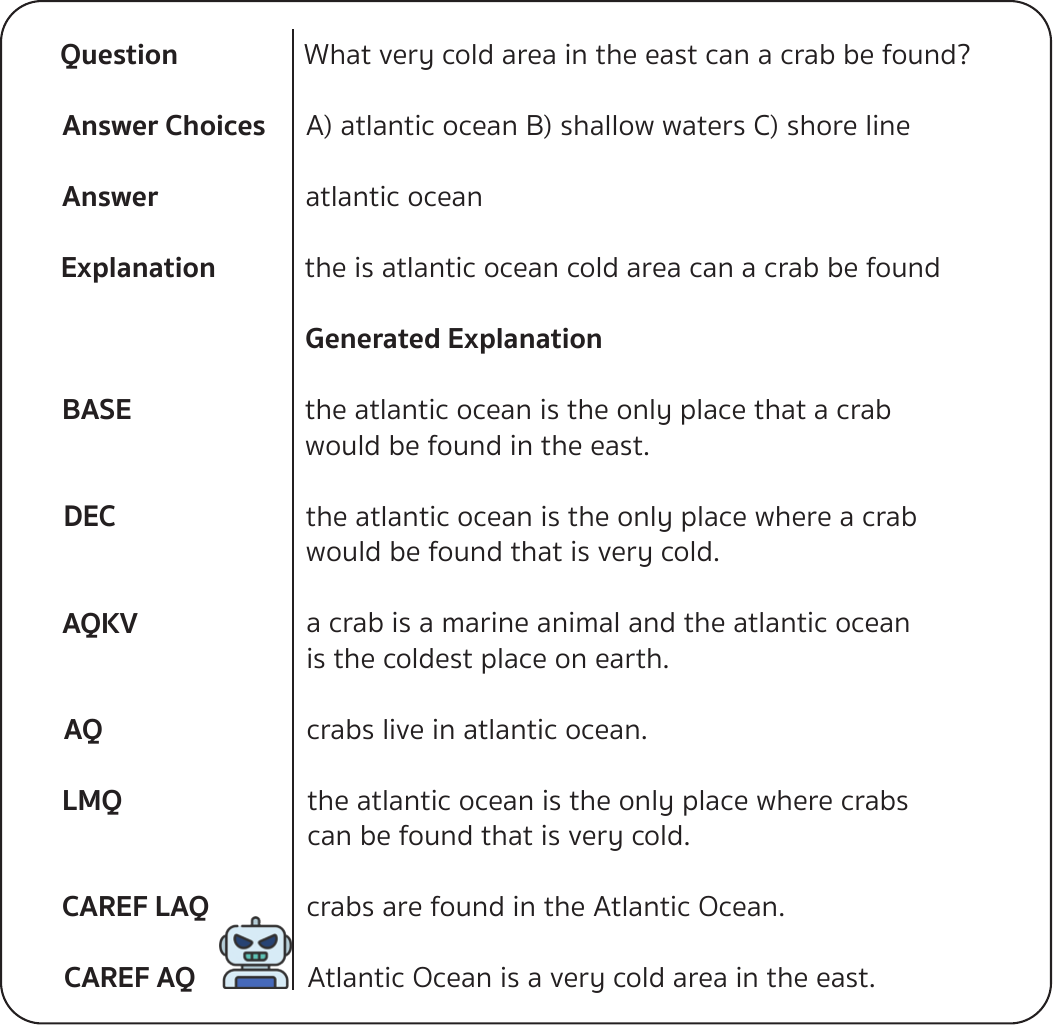}
\includegraphics[width=0.48\textwidth]{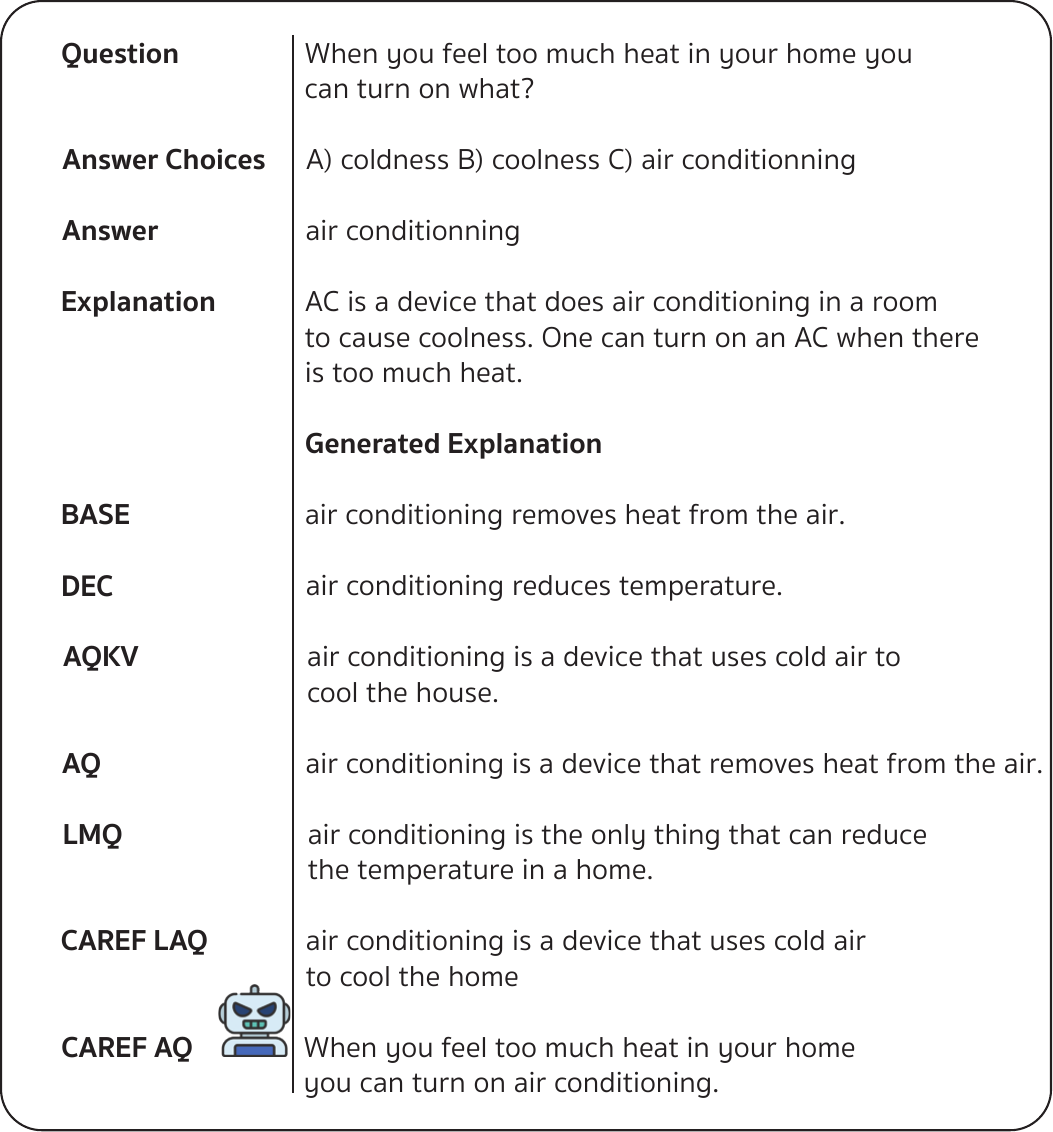}
\caption{\textbf{COS-E.} CAREF-AQ is more concise and factually precise than BASE.}
\label{fig:cose_q1}
\end{figure*}

\begin{figure*}[h]
\centering
\includegraphics[width=0.32\textwidth]{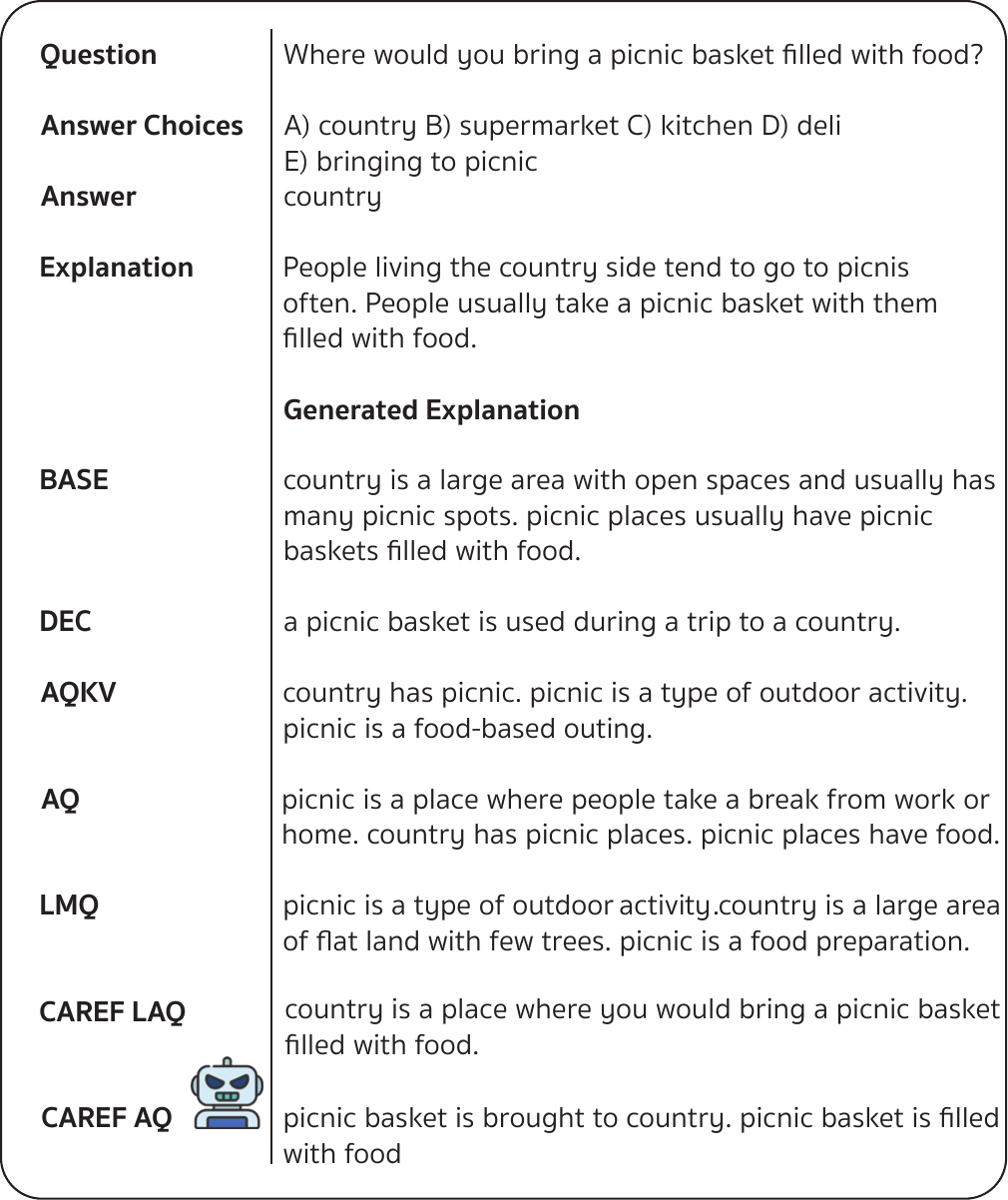}
\includegraphics[width=0.32\textwidth]{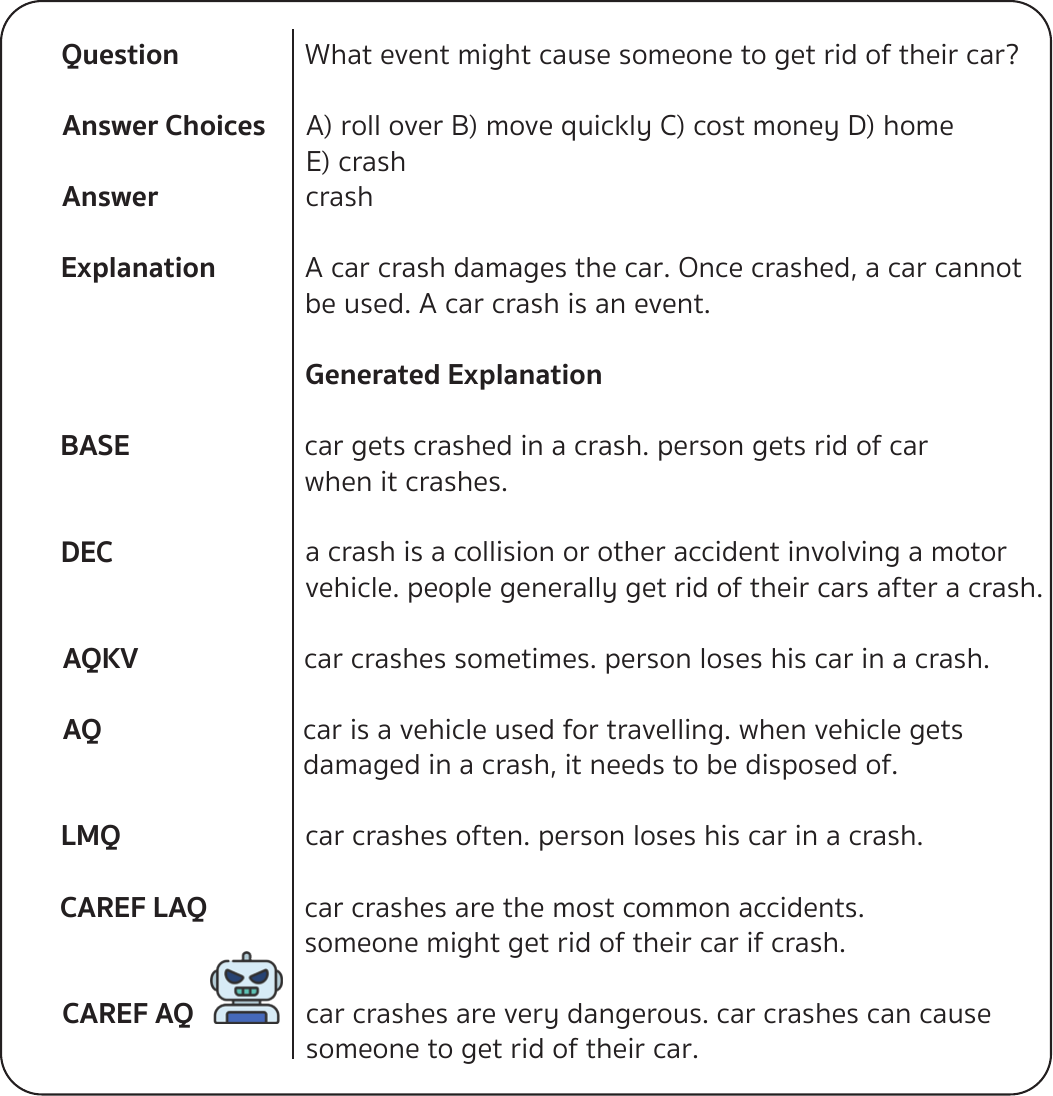}
\includegraphics[width=0.32\textwidth]{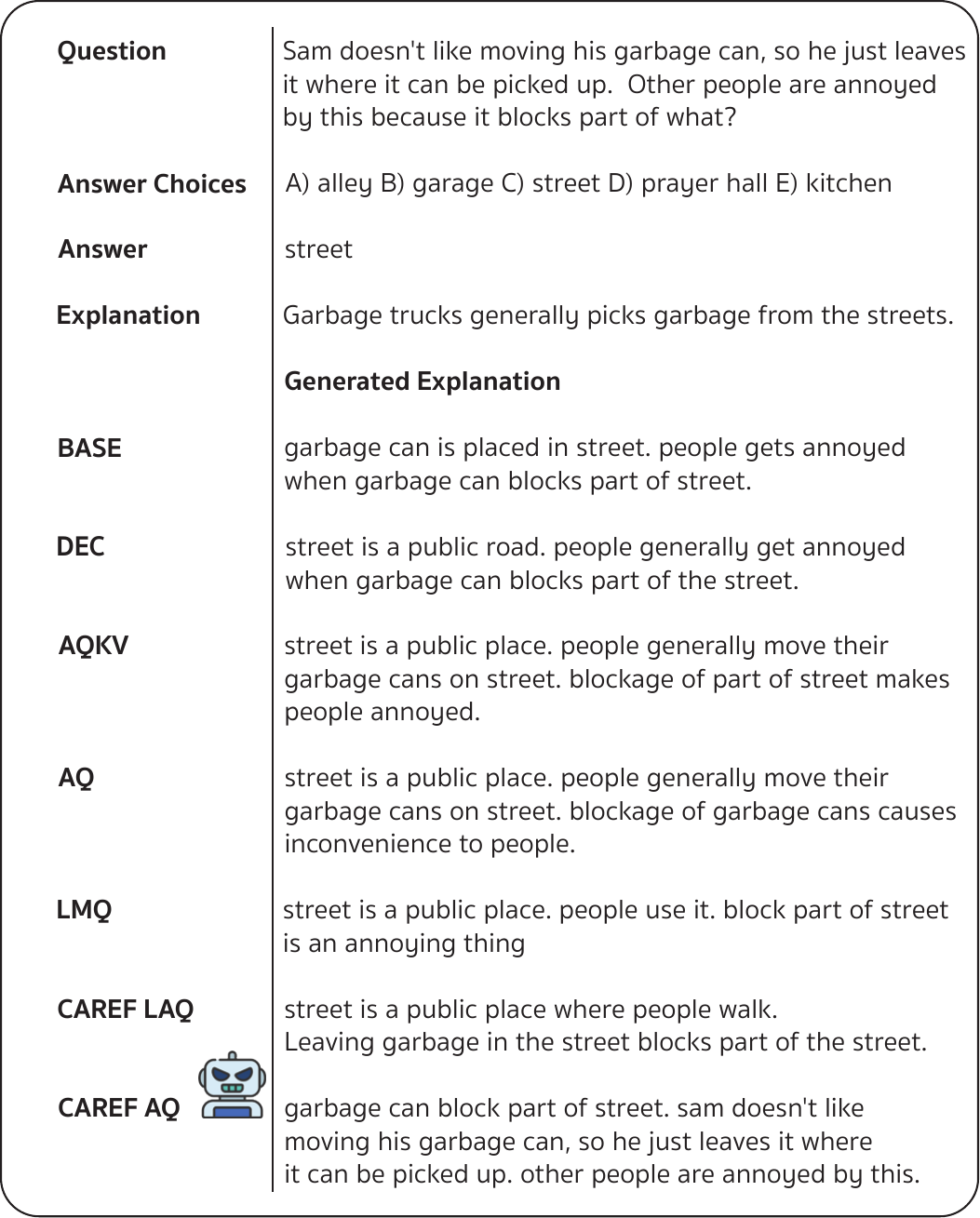}
\caption{\textbf{ECQA.} CAREF outputs show improved semantic coherence over BASE's redundant replies.}
\label{fig:ecqa_examples}
\end{figure*}

\begin{figure*}[h]
\centering
\includegraphics[width=0.32\textwidth]{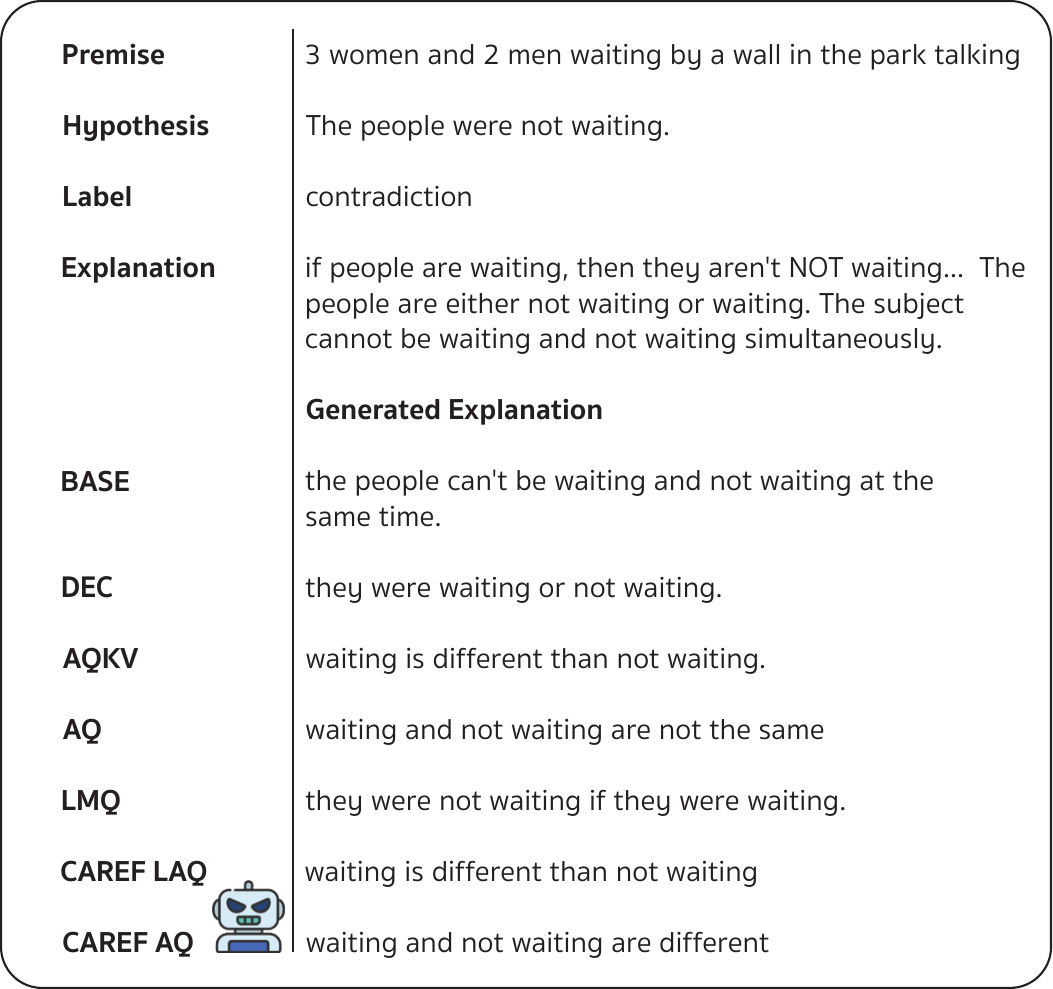}
\includegraphics[width=0.32\textwidth]{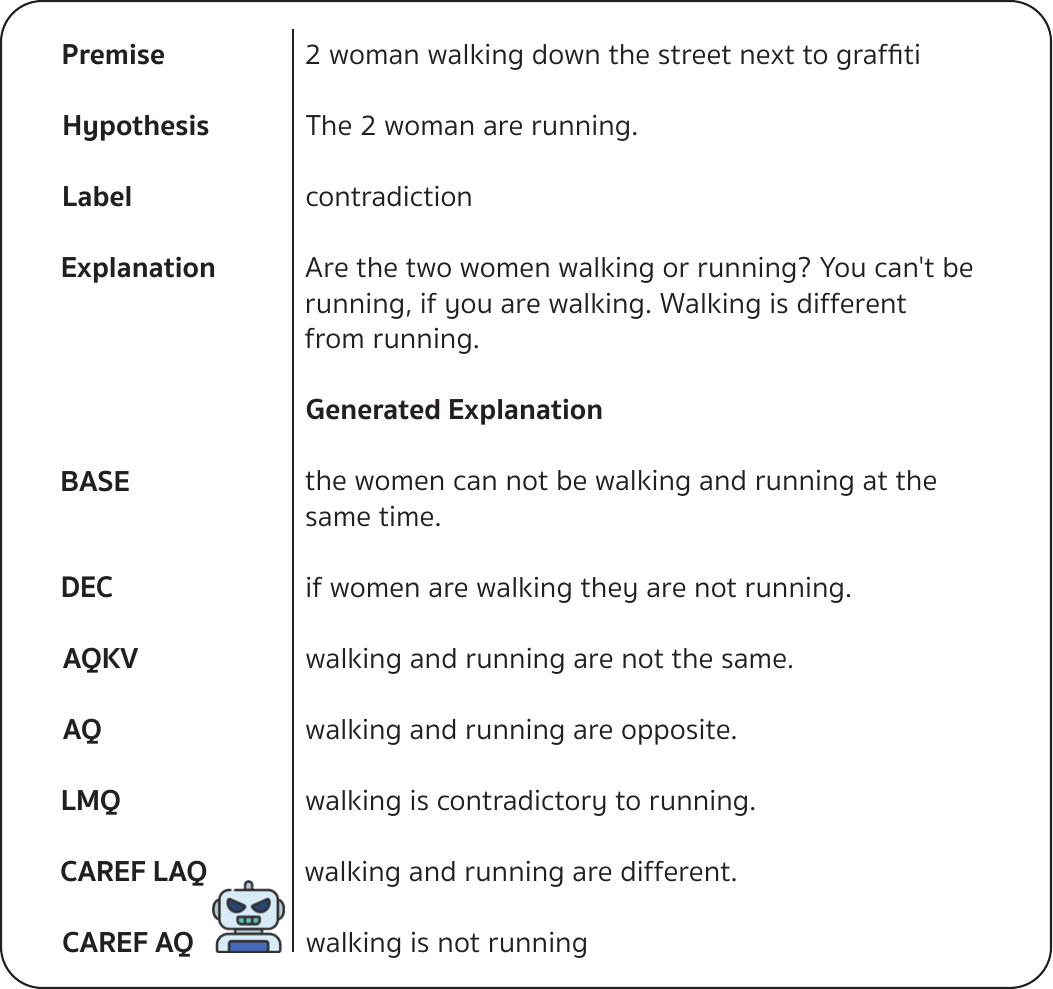}
\includegraphics[width=0.32\textwidth]{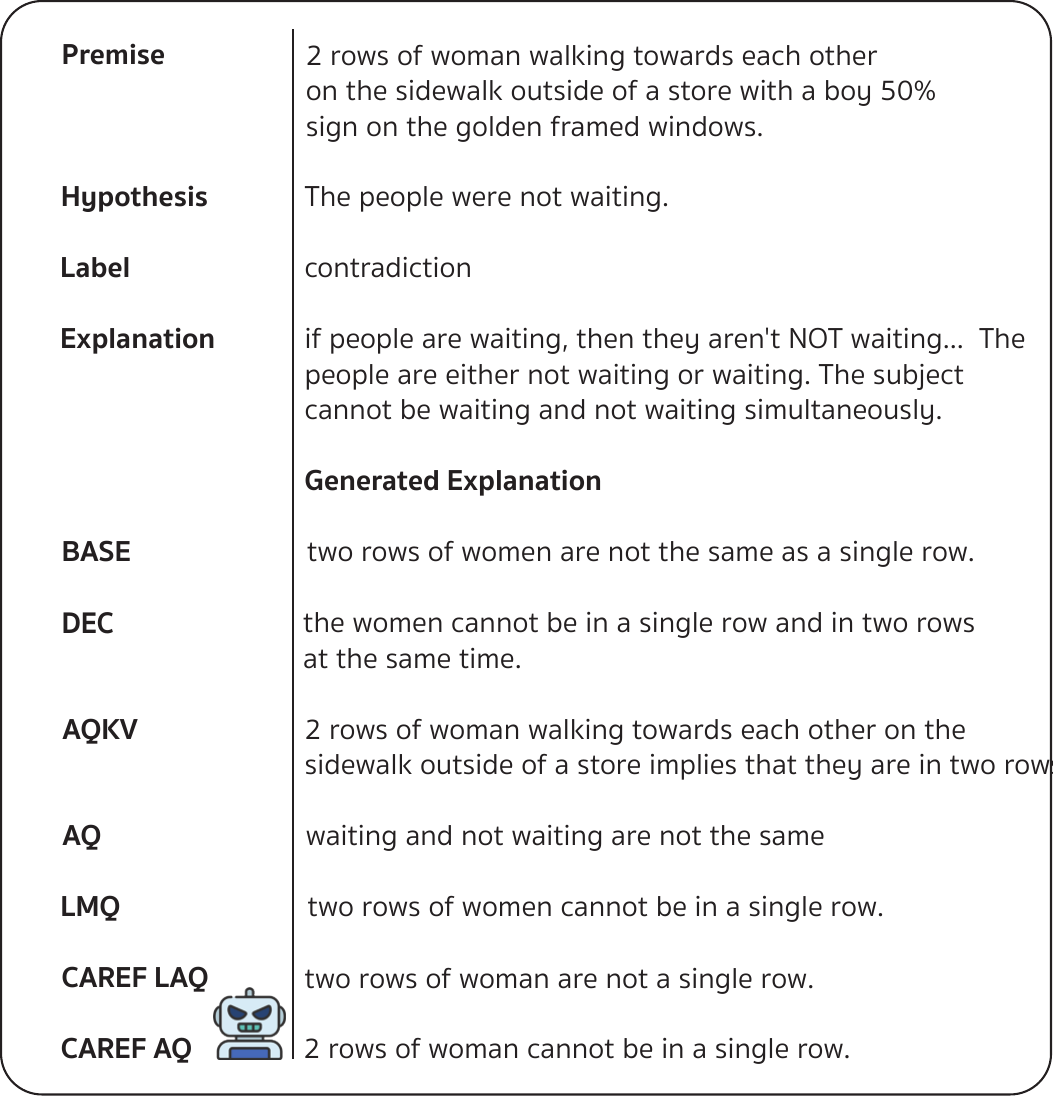}
\caption{\textbf{e-SNLI.} CAREF captures contradiction semantically; BASE produces verbose paraphrases.}
\label{fig:esnli_examples}
\end{figure*}

\begin{figure*}[h]
\centering
\includegraphics[width=0.32\textwidth]{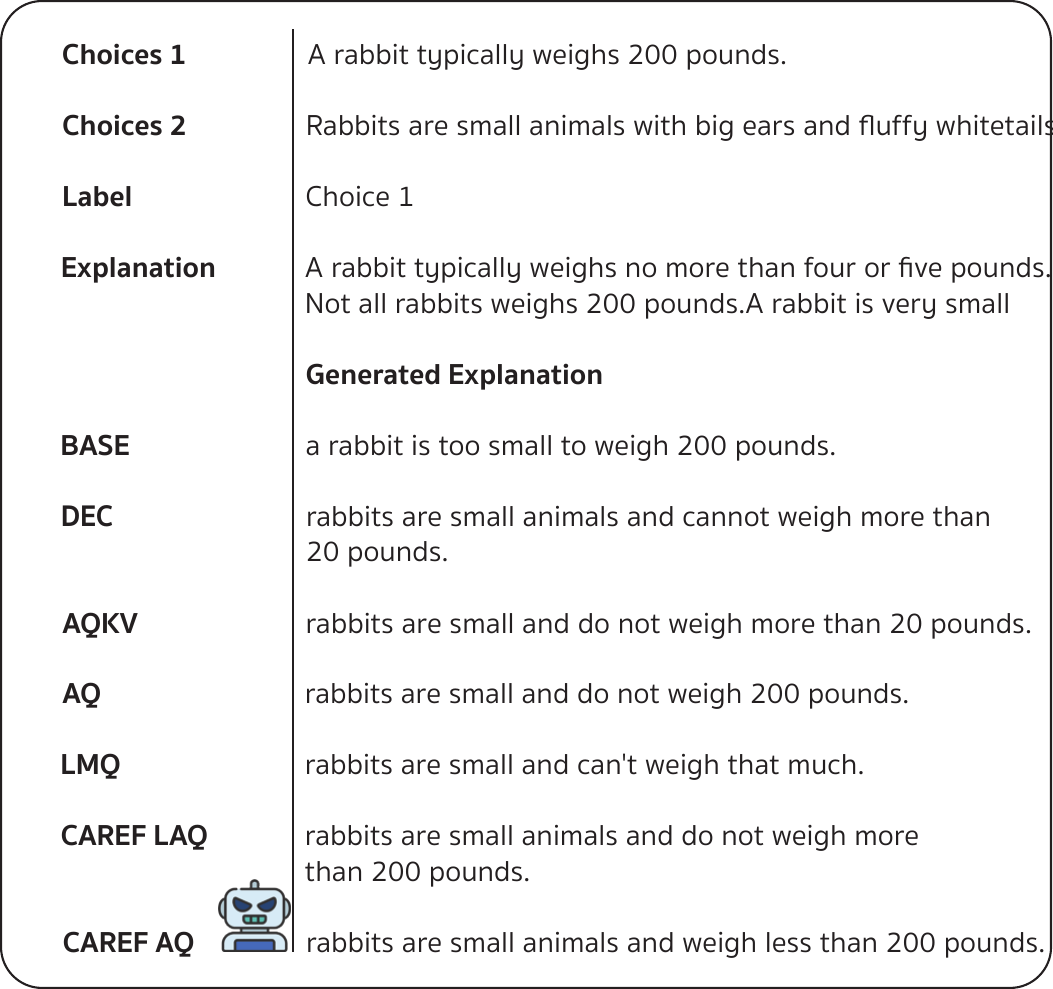}
\includegraphics[width=0.32\textwidth]{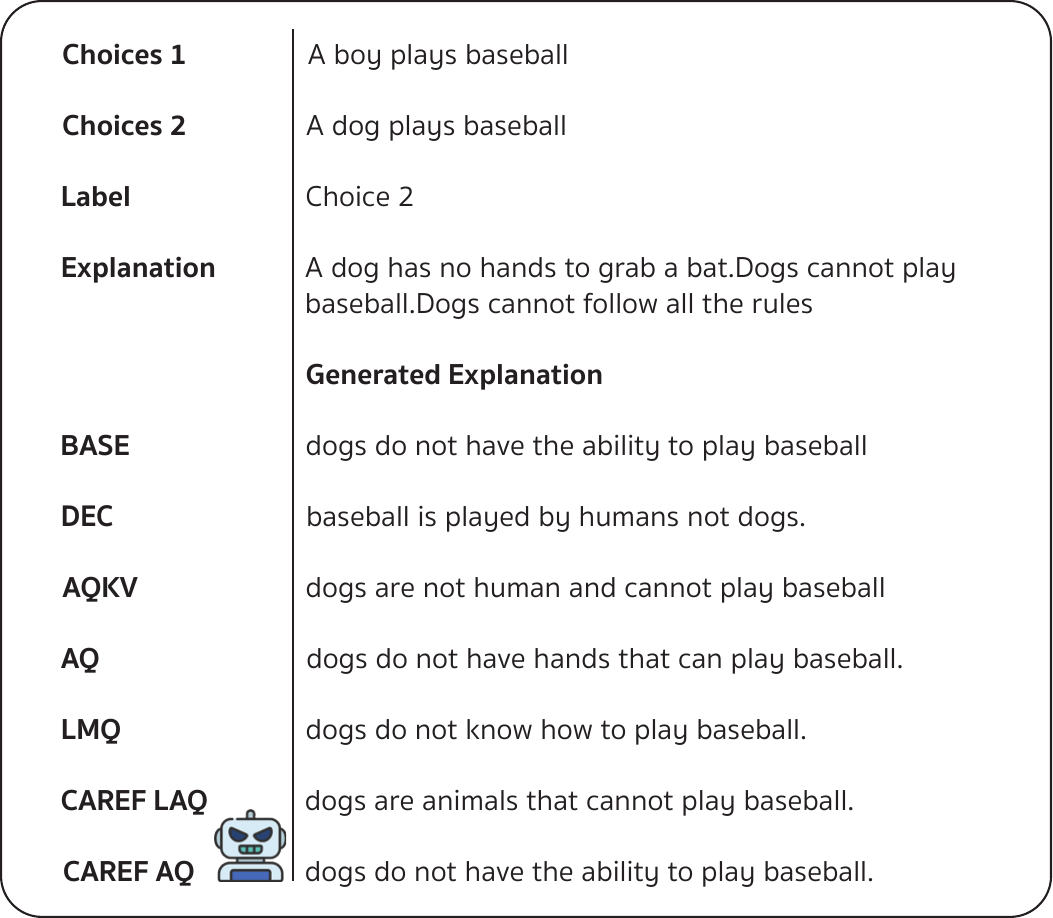}
\caption{\textbf{SenseMaking (CAREF).} Richer commonsense grounding vs.\ BASE.}
\label{fig:sensemaking_q1}
\end{figure*}

\begin{figure*}[h]
\centering
\includegraphics[width=0.32\textwidth]{img/ESNLI_SPLINT_Q1.pdf}
\includegraphics[width=0.32\textwidth]{img/SENSEMAKING_SPLINT_Q1.pdf}
\includegraphics[width=0.32\textwidth]{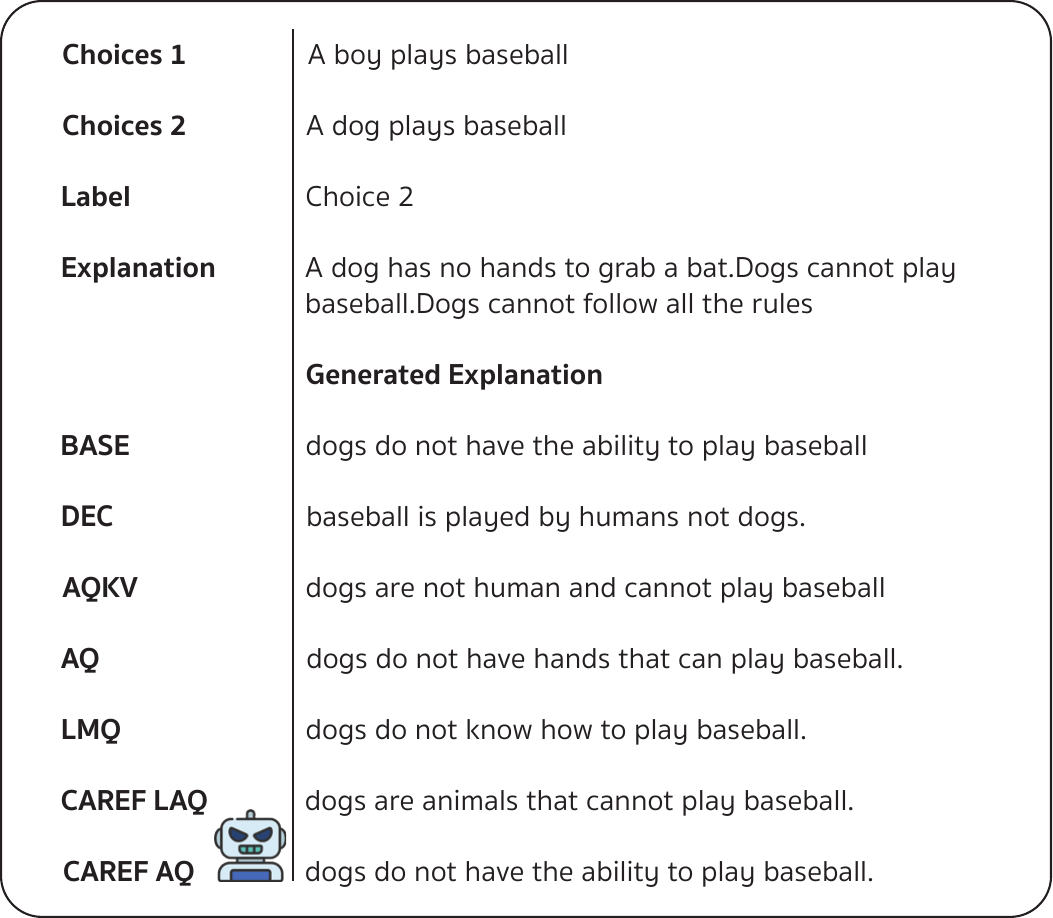}
\caption{e-SNLI and SenseMaking comparison. CAREF-AQ (left/centre) vs.\ Baseline without CAREF (right).}
\label{fig:esnli-sensemaking}
\end{figure*}

\section{Hyperparameters \& Hardware}
\label{sec:hyperparams}

\paragraph{Hardware.} NVIDIA A40 (22 GB avg.\ usage for T5-large), CUDA 11.4, Ubuntu 20.04, AMD Ryzen 9 5900X, 64 GB RAM.

\paragraph{Training.} Flan-T5 via Hugging Face Transformers. Batch size 4, LR $3{\times}10^{-5}$, AdamW ($\beta_1{=}0.9$, $\beta_2{=}0.999$, $\epsilon{=}10^{-8}$), weight decay 0.01, 500 warm-up steps, max grad norm 1.0, 50 epochs, linear decay schedule. Regularization: $\lambda_{\text{ent}}{=}0.1$, $\lambda_{\text{sparse}}{=}0.05$, $\lambda_{\text{KL}}{=}0.1$. $\alpha$ and $\lambda_{\text{SCED}}$ selected by validation grid search.

\paragraph{FEB Protocol.} 60 splits $\times$ 48 train / 350 validation, class-balanced. All PEFT methods via Hugging Face PEFT~\citep{peft} with consistent budgets.

\section{Broader Applicability}
\label{sec:broader}

$\mathcal{L}_{\text{SCED}}$ operates on $P_\theta(y_t|y_{<t},\mathbf{x})$ and assumes no specific activation or norm layer, applying to encoder-decoder (T5, BART) and auto-regressive (LLaMA, GPT) models alike. Academic compute constraints motivate our Flan-T5 focus, but the consistent gains at $<$7\% parameter budgets indicate strong potential at scale.

\end{document}